\documentclass[letterpaper, 10 pt, conference]{ieeeconf}  

\IEEEoverridecommandlockouts                              

\usepackage{graphics} 
\usepackage{epsfig} 
\usepackage{mathptmx} 
\usepackage{times} 
\usepackage{amsmath} 
\usepackage{amssymb}  
\usepackage{algorithm}
\usepackage{algpseudocode}
\usepackage{tabularx}
\usepackage{xcolor}
\usepackage{subcaption}
\usepackage{graphicx}
\usepackage{balance}

\title{\LARGE \bf
Feelit: Combining Compliant Shape Displays with Vision-Based Tactile Sensors for Real-Time Teletaction
}

\author{Oscar Yu$^{1}$ and Yu She$^{2,*}$
\thanks{*Corresponding Author}
\thanks{$^{1}$Oscar Yu is with the School of Electrical \& Computer Engineering, Purdue University
        West Lafayette, IN 47907, USA
        {\tt\small oyu@purdue.edu}}%
\thanks{$^{2}$Yu She is with the School of Industrial Engineering, Purdue University
        West Lafayette, IN 47907, USA
        {\tt\small shey@purdue.edu}.} 
}

\begin{document}

\maketitle
\thispagestyle{empty}
\pagestyle{empty}

\begin{abstract}

Teletaction, the transmission of tactile feedback or touch, is a crucial aspect in the field of teleoperation. High-quality teletaction feedback allows users to remotely manipulate objects and increase the quality of the human-machine interface between the operator and the robot, making complex manipulation tasks possible. Advances in the field of teletaction for teleoperation however, have yet to make full use of the high-resolution 3D data provided by modern vision-based tactile sensors. Existing solutions for teletaction lack in one or more areas of form or function, such as fidelity or hardware footprint. In this paper, we showcase our design for a low-cost teletaction device that can utilize real-time high-resolution tactile information from vision-based tactile sensors, through both physical 3D surface reconstruction and shear displacement. We present our device, the Feelit, which uses a combination of a pin-based shape display and compliant mechanisms to accomplish this task. The pin-based shape display utilizes an array of 24 servomotors with miniature Bowden cables, giving the device a resolution of 6x4 pins in a 15x10 mm display footprint. Each pin can actuate up to 3 mm in 200 ms, while providing ~80 N of force and 1.5 um of depth resolution. Shear displacement and rotation is achieved using a compliant mechanism design, allowing a minimum of 1 mm displacement laterally and 10 degrees of rotation. This real-time 3D tactile reconstruction is achieved with the use of a vision-based tactile sensor, the GelSight \cite{yuan2017gelsight}, along with an algorithm that samples the depth data and marker tracking to generate actuator commands. Through a series of experiments including shape recognition and relative weight identification, we show that our device has the potential to expand teletaction capabilities in the teleoperation space.

\end{abstract}

\section{INTRODUCTION}

Robotic Teleoperation is defined as the process in which a human operator, through the use of a human-machine interface, remotely operates a robotic mechanism, allowing them to manipulate and sense objects through feedback of the human-machine interface. Robotic teleoperation technologies have found many uses outside of the research space \cite{cui2003review}. Medical professionals utilize surgical teleoperation robots to perform surgeries remotely and access hard-to-reach areas of the body, minimizing invasiveness. In austere and hazardous environments such as the reactor cores of nuclear plants or the vacuum of space, teleoperation is a vital tool that keeps human operators safe when carrying out critical tasks. 

\begin{figure}[t]
      \centering
      \includegraphics[scale=0.17]{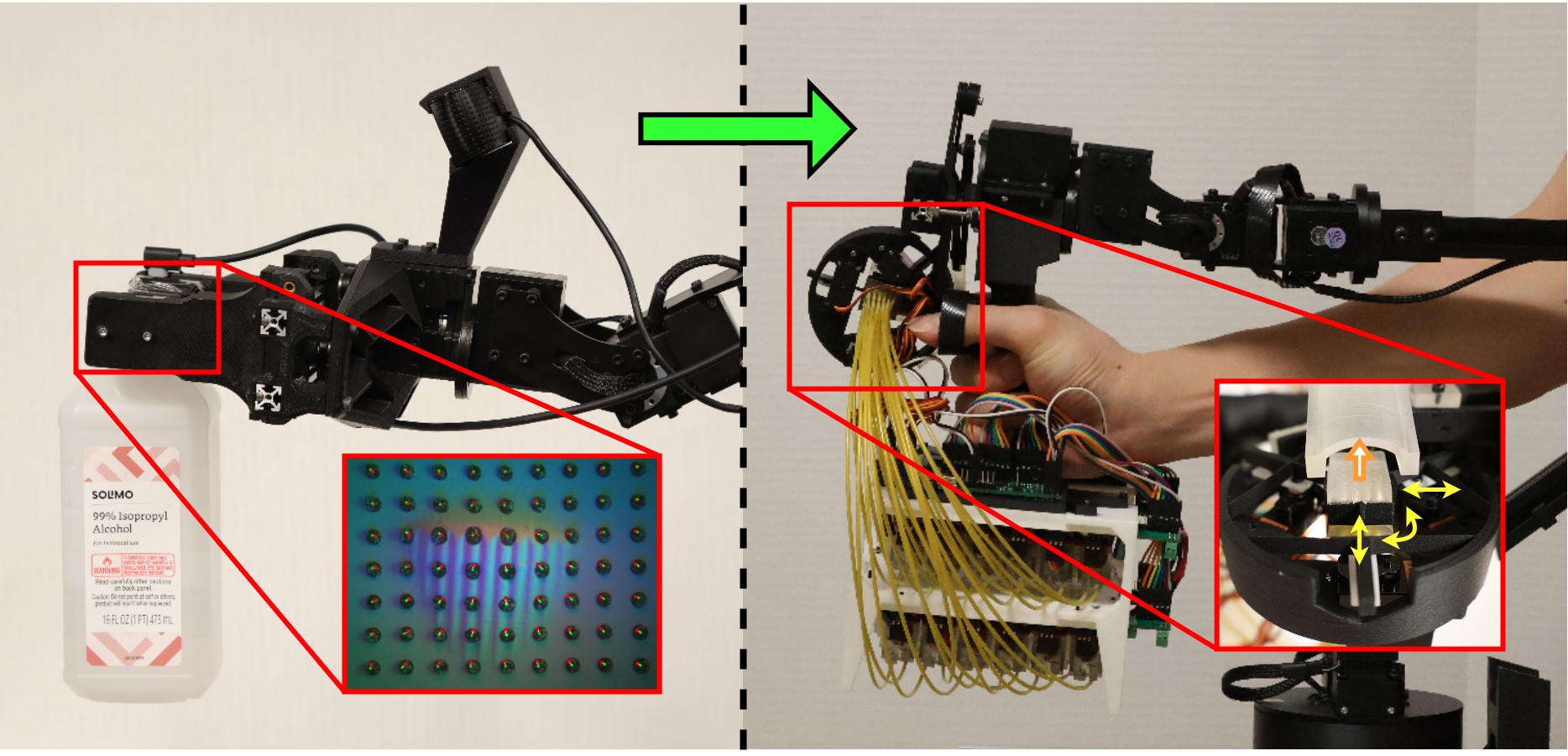}
      \caption{The teletaction device physically reconstructs depth and shear information provided by the GelSight vision-based tactile sensor in real-time.}
      \label{fig:overview}
\end{figure}

In the field of robotic teleoperation, there is the concept of `telepresence' in which the human operator sufficiently feels as if they themselves are present where the robot is operating; in other words, the quality of the human-machine interface and the feedback presented are enough to recreate the operational environment to a certain degree \cite{cui2003review}. Forms of feedback include but are not limited to simple force-feedback transducers from the end effector, Augmented Reality integration, and more. Specifically, our research is focused in `teletaction', a facet of telepresence, that encompasses the transmission of cutaneous information of a surface such as shape or texture to an operator's skin \cite{606758}.
Research into human sensory perception show that the utility of high-quality tactile feedback holds promise, especially in regard to dexterous manipulation and tasks requiring sensing of minute features \cite{6788048}.
In tasks requiring rich tactile feedback, not only is depth information useful, but shear forces and displacements as well. Humans are able to sense tangential shear forces using a variety of receptors in the skin. Psychophysics research has shown that pattern discrimination and sensitivity to stimuli increase when stimulated at different times, increasing spatial resolution \cite{tacperception}. 

The inception of vision-based tactile sensors represent a jump in tactile sensing capabilities for robotics. These sensors triumph over traditional methods such as resistive, capacitive, or piezoelectric sensors; as they provide much higher spatial resolution, and can supply additional information such as shear forces and contact geometry \cite{DBLP:journals/corr/abs-2106-08851}. These sensors such as the GelSight \cite{yuan2017gelsight}\cite{wang2021gelsight} have been used to perform dexterous manipulation tasks such as cable following \cite{DBLP:journals/corr/abs-1910-02860}. 

Development and application of teletaction systems have also yielded improved human performance and sensing when used with telepresence systems. However, most developments in the teletaction space have their share of drawbacks as well. Some compact wearable systems, such as the Haptic Thimble \cite{7463168} and others \cite{10.1145/2701973.2702719}\cite{Pierce2014AWD} are not able to transmit high-fidelity surface information, relying on other qualities like force feedback or vibration for teletaction. Recently, Carnegie Mellon's Future Interface Systems group has developed the Fluid Reality haptic fingertip\cite{10.1145/3586183.3606771}, which currently represents the smallest form factor of wearable haptic devices that are able to transmit depth and contact information. Even still, this solution lacks in tactile resolution, utilizing only two levels of depth actuation per pin. Other solutions are able to emulate other facets of telepresence, such as shear forces or stretchable surfaces, but have a more niche application space or a bulky footprint that would be difficult to integrate with telepresence systems \cite{10.1145/2701973.2702719}\cite{Steed_Ofek_Sinclair_Gonzalez-Franco_2021}\cite{9812131}.

\begin{figure}[t]
      \centering
      \includegraphics[scale=0.27]{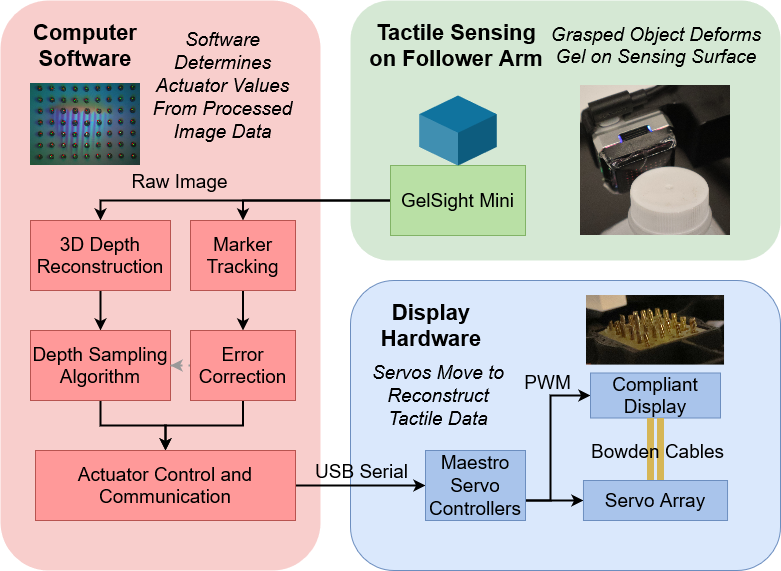}
      \caption{Component overview of the teletaction device.}
      \label{fig:blockdiagram}
\end{figure}

The current most popular method of recreating surface geometry to a high degree of resolution is the pin-based shape display. This method utilizes multiple actuators to actuate a grid of pins up or down, in which the height of the pins correspond to a particular surface geometry. Past research into pin-based shape displays include MIT's inFORCE table and TRANS-DOCK system \cite{10.1145/2501988.2502032}\cite{10.1145/3374920.3374933} and FEELEX \cite{10.1145/383259.383314}. Designs utilizing lower cost commercially available components have been demonstrated as well \cite{998981}.

Pin based shape displays also have their own drawbacks. Most designs typically exhibit a large footprint, inhibiting their use with telepresence systems. Those utilizing pneumatic systems \cite{845247} or electromagnetic braking \cite{10.1145/2858036.2858264} that can combine both wearability and fidelity have reduced refresh rates and low actuator force, limiting their application.

Our research focuses on the development of a miniaturized, low-cost pin based shape display device with compliant mechanisms to leverage the full capabilities offered by vision-based tactile sensors when used in robotic telepresence applications. Our contribution to the teleoperation space is our device, which we call Feelit, can perform both \textbf{physical 3D depth reconstruction} and \textbf{physical shear displacement reconstruction} in \textbf{real-time}, and demonstrate it's advantages in \textbf{teleoperation} by performing psychophysics and teleoperation experiments, including \textbf{shape recognition} and \textbf{relative weight identification}.

\section{Design Overview}

\begin{figure}[t]
      \centering
      \includegraphics[scale=0.23]{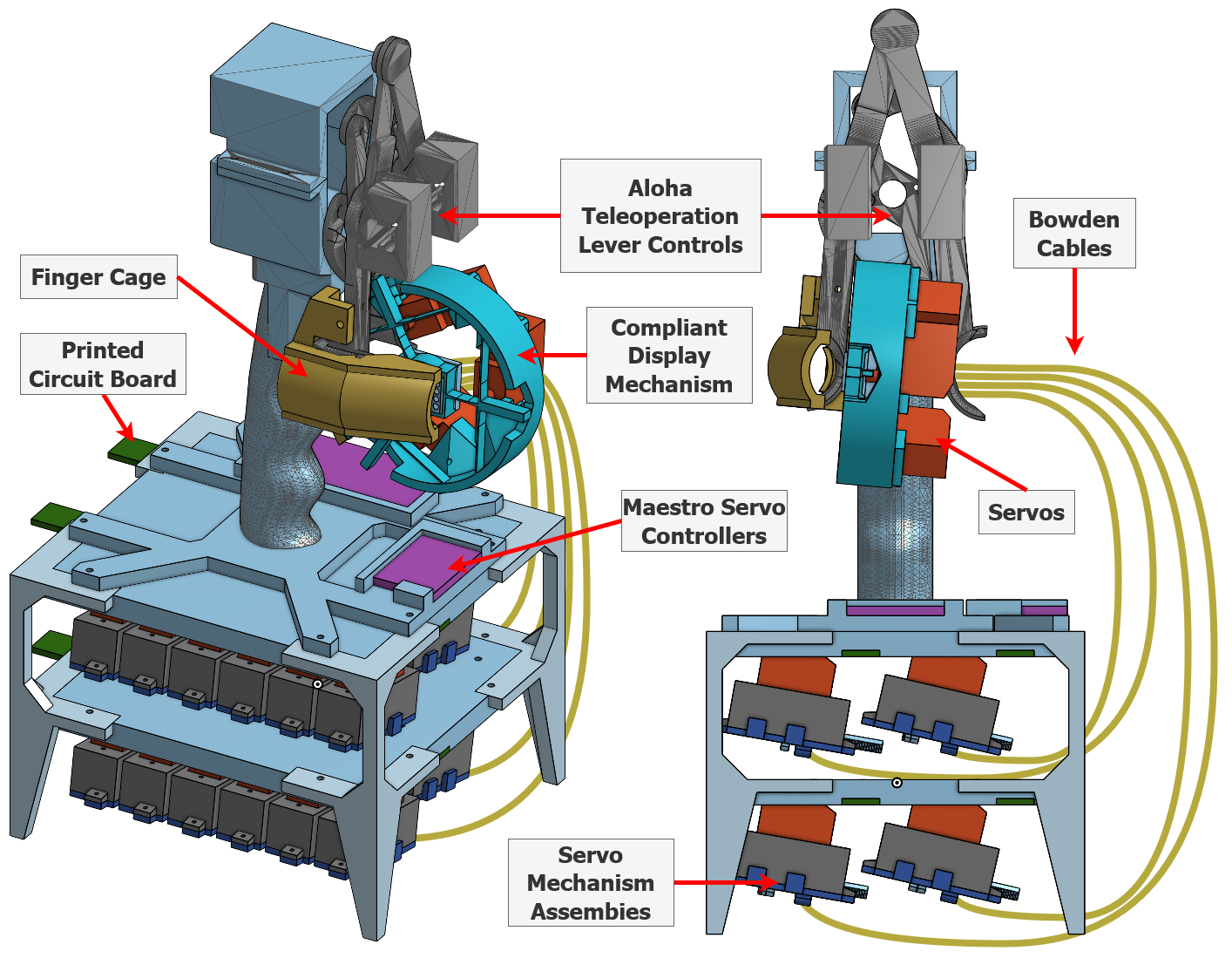}
      \caption{Component overview of the teletaction device.}
      \label{fig:blockdiagram}
\end{figure}


Fig. \textcolor{red}{2} shows a block diagram overview of the Feelit tactile reconstruction device, which is mounted on the leader arm of the Aloha teleoperation system \cite{zhao2023learning}. The Aloha teleoperation system consists of two ViperX 6-DoF robot arms with bimanual parallel-jaw grippers attached to the end effectors. The leader arm is controller by the user, which the follower arm mimics. First, the Aloha follower arm grasps an objects with the GelSight tactile sensor attached to the end effector. The GelSight captures live video of the deformed gel, and transmits the video to the computer over USB. On the computer, the video stream is processed to reconstruct the 3D depth and shear displacement information. An algorithm then determines the actuator commands in terms of Pulse Width Modulation (PWM) duty cycles needed to reconstruct the tactile information. The user can specify options to control the scale of reconstruction, as well as the size and location for sampling area. The individual servo actuation data is communicated over serial USB to the Mini Maestro Servo Controllers, which control of individual servos in the tactile device. On the pin display, the servos are then actuated, applying force onto the Bowden cables attached to the display face. This deforms the silicon gel on which the cable ends are cured to (not shown in the figure). The compliant mechanism, which controls the position of the display face, also has servos for actuation. The result is a real-time reconstruction of the 3D surface and shear displacements sensed by the GelSight onto the pin display face. The form factor for our device is made specifically to interface with the Aloha Teleoperation system, with the compliant mechanism and display face attached to the manipulator controls, and the cables leading to servo housing placed under the grip. The material costs of our device total less than \$500, making it economically competitive with other teletaction solutions. A physical layout of the device is shown in Fig. \textcolor{red}{3}.

\subsection{Actuator Design \& Bowden Cables}

Different actuation methods were considered, including standard digital servos, pneumatic actuators \cite{845247}, electrostatic braking pins \cite{8845668}, and linear actuators. For our design we opted to use standard commercial-off-the-shelf servo motors, which are EMAX ES08MA II 12g Metal Gear servos, along with the Pololu \emph{Mini-Maestro 24 Channel USB servo controller}, which serves as an interface between the software and the servos. A rack-and-pinion design was employed for the actuation mechanism. The range of motion for the actuation mechanism is 3 mm, with 1.5 um of resolution. The output torque was calculated at 78 N and actuation speed was measured at 15 mm/s. 

We use Bowden cables, specifically the Gold-n-Rod \emph{Pushrod System, 36" Cable .032"}, to transfer force from the actuators to the display face. These cables are small in diameter and are flexible, allowing for the display face to move for shear displacement reconstruction and for the actuator bodies to be located away from the display face.

\begin{figure}[t]
      \centering
      \begin{subfigure}[t]{0.18\textwidth}
            \centering
            \includegraphics[width=\textwidth]{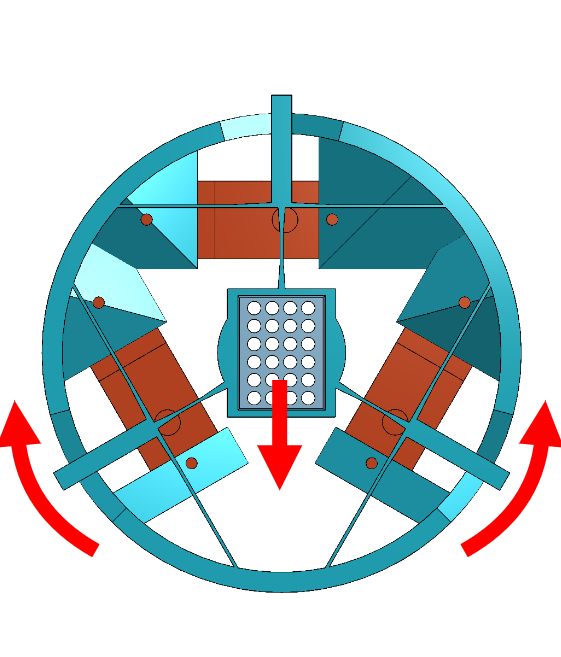}
            \subcaption{ Lateral movement}
      \end{subfigure}
      \hfill
      \begin{subfigure}[t]{0.18\textwidth}
            \centering
            \includegraphics[width=\textwidth]{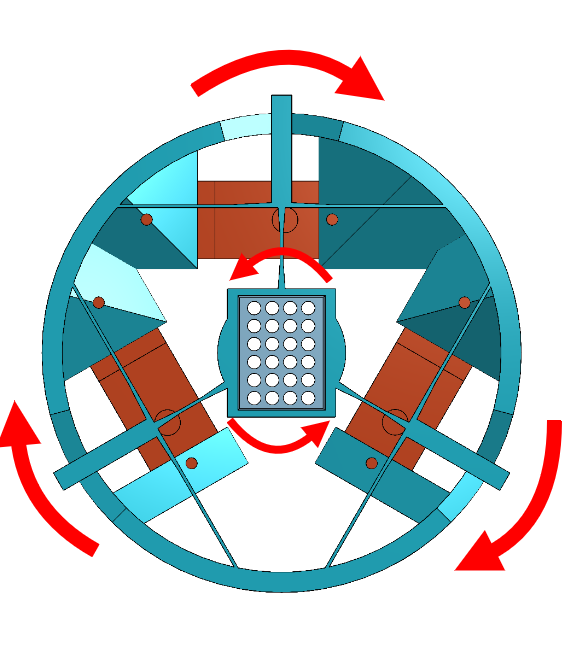}
            \subcaption{ Rotation}
      \end{subfigure}
      \caption{Examples of stage movements as a result of tab actuation. Deflection of the tabs flexes the joints connected to the display stage, causing it to move.}
\end{figure}

\subsection{Compliant Display Face}

The display face is a resin-printed box with a grid of 6x4 holes, of which pins formed by the ends of the Bowden cables extend out of. The pins are separated by a pitch of ~2.6 mm, giving the display a size of 15 mm by 10 mm, approximately the size of a human finger. A 3mm layer of silicone elastomer, similar to the GelSight face, is cured to the pins to give the user continuous surface to feel.

The display face itself is mounted onto a compliant mechanism which is a modified design based off the HexFlex \cite{hexflex} compliant stage. The compliant stage has 3 degrees of freedom: lateral movement and rotation tangential to the display face. Actuation is achieved by rotating the 3 outer tabs, the combination of which contributes to the displacement of the stage. The three outer tabs are connected to specialized servo horns that allow the tabs to rotate even when displaced. Fig. \textcolor{red}{4} gives a graphical example of stage movements as a result of tab actuation. 

An accurate kinematic model is needed to precisely actuate the display; however analytical modeling of compliant mechanisms are notoriously complex. We opted instead to estimate the kinematics to a reasonable degree of accuracy (\textless0.1 mm) for our application. We assume that there is a linearly independent relationship between the displacement of the compliant stage, and the actuation of each tab. Each tab actuation contributes to the lateral and rotational displacement of the compliant stage, which can be modeled with an nth degree polynomial. The contributions of each tab are then combined to find the total displacement.

\begin{figure}[t]
      \centering
      \begin{subfigure}[t]{0.4\textwidth}
            \centering
            \includegraphics[width=\textwidth]{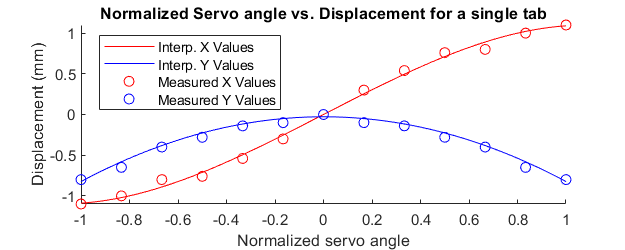}
            \subcaption{Lateral movement}
      \end{subfigure}
      \hfill
      \begin{subfigure}[t]{0.4\textwidth}
            \centering
            \includegraphics[width=\textwidth]{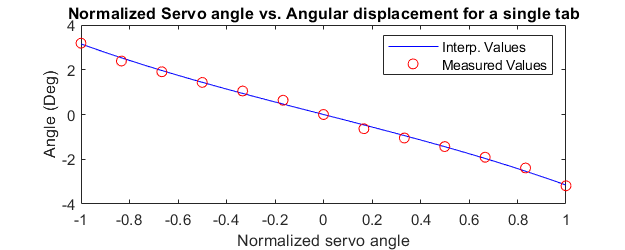}
            \subcaption{Rotation}
      \end{subfigure}
      \caption{Measured and interpolated values for stage lateral \& rotational displacement vs. single servo actuation.}
\end{figure}

\[
\begin{bmatrix}
x\\
y\\
\phi\\
\end{bmatrix} 	
=
\begin{bmatrix}
p_{x_1}(\theta_1) + p_{x_2}(\theta_2) + p_{x_3}(\theta_3)\\
p_{y_1}(\theta_1) + p_{y_2}(\theta_2) + p_{y_3}(\theta_3)\\
p_{\phi_1}(\theta_1) + p_{\phi_2}(\theta_2) + p_{\phi_3}(\theta_3)\\
\end{bmatrix} 	
\]
where
\[p_(x) = p_1x^n + p_2x^{n-1} + ... + p_{n-1}x + p_n\]
and
\[-1 \leq \theta_{1,2,3}\ \leq 1\] are the normalized servo angles for each actuation tab 

This expression is then used to find the Jacobian, which is used in a vanilla iterative inverse kinematics algorithm to find the servo angles for a desired \(x,y,\) and \(\phi\). From test bench measurements (Fig. \textcolor{red}{5}), the compliant display has a minimum of 1mm of lateral movement in any direction without rotation, and 10 degrees of rotation about the center without any lateral movement, as one affects the range of the other.

\subsection{Depth Reconstruction}

To understand the advantage vision-based tactile sensing give over traditional tactile sensors such as capacitave sensors, A brief explanation of the sensor is needed. Specifically, the sensor used in this research is the GelSight. The sensor consists of a silicone elastomer face coated with a gray or silver specular paint, which forms tactile sensing area. An array of red, green, and blue LEDs illuminate the elastomer from different directions parallel to the face. When the elastomer comes into contact with an object it deforms, causing the lights to illuminate its surface, which is captured by a camera. This image is then used to estimate the depth map using Photometric Stereo technique \cite{photostereo}, that creates a height function \(z_i = f(x_i,y_i)\) which maps a pixel location \(x_i\) and \(y_i\) to a depth \(z_i\). A quick overview of the Photometric Stereo technique is as follows: First, a neural network is used to learn a color-to-gradient mapping of each R, G, B value at \((x, y)\) to its corresponding deformed surface gradient \((G_x^i,G_y^i)\). This is done using impressions of a 3D printed sphere of known diameter onto the gel, and assigning surface gradients manually, to use as training data. After the neural network learns the color-to-gradient mapping, a 2-D fast Possion solver is used to spatially integrate the surface gradients to obtain the depth map. The result is a depth map of the elastomer surface deformed by the object, with a resolution of 240 by 320 pixels over a 18 by 24 mm sensing area. Vision based tactile sensors are able to sense much finer details in depth and texture compared to traditional contact sensors, and estimate shear displacement forces using markers on the silicone face \cite{yuan2017gelsight}.

\subsection{Marker Tracking \& Correction}

To provide shear displacement information, a grid of markers are needed on the gel face of the GelSight sensor. Accurate marker tracking is needed for shear displacement calculation, and to provide masking information for the depth estimation, as the different color of the markers interfere with the depth calculation algorithm. Tracking is performed with the mean-shift algorithm provided in the codebase for the GelSight sensor. Mean shift was chosen as it is less prone to drift and edge case errors compared to optical flow tracking, and is comparable in computational speed with mean-shift operating at an average of 16.76 frames per second (FPS) and optical flow with 15.95 FPS over 500 frames. The mean-shift algorithm iteratively moves each tracked marker to the nearest high-density region in an altered grayscale image. True marker locations in the input image are represented as high-density regions, and ideally the last known tracked marker locations are locally near their corresponding true marker, allowing the algorithm to converge. 

The mean-shift algorithm is still prone to errors during transient moments of high displacement, with markers either losing track or converging onto each other. To combat this, a marker correction algorithm is implemented to identify \& correct untrustworthy markers. The psuedocode for the marker error correction algorithm is given in \textbf{Algorithm 1}

\begin{algorithm}
\caption{Marker Error Correction}\label{alg:cap}
\begin{algorithmic}
\State $img \gets GelSight\_Live()$ 
\State $bin\_img \gets cv2.adaptiveThreshold(cv2.cvtColor(img)))$  
\State $marker\_list \gets mean\_shift\_alg(img)$
\State $trust\_idx \gets np.ones()$
\For{$i \in length(marker\_list)$}
\State{$curr\_marker \gets marker\_list(i)$}
\If{$curr\_marker$ $location$ in $bin\_img$ is $1$ \\ \textbf{or} $\mid curr\_marker.vec \mid > 30$}
    \State $trust\_idx(i) \gets 0$
\EndIf
\For{$adj\_marker \in surrounding\_markers$}
\State{$d \gets \mid curr\_marker.vec - adj\_marker.vec\mid$}
\If{$d > max\_diff$}
    \State $trust\_idx(i) \gets 0$
    \State \textbf{break}
\EndIf
\EndFor
\EndFor
\For{$j \in trust\_idx = 0$}
\State{$marker\_list(j) \gets interpolate(marker\_list(trust\_idx))$}
\EndFor
\end{algorithmic}
\end{algorithm}

We first parse through each estimated marker and determine trustworthiness with several critera. First, we check is if the current evaluated marker is located on a dark spot in the image, which generally corresponds to marker locations. This is done by transforming the image into grayscale, then to a binary image via dynamic thresholding. Dark pixels in the image are represented as zeros, and generally correspond to marker shapes. We cannot utilize this binary image as a mask directly as large displacements are also present in the binary image. Next, we check if the vector displacement of the current marker generally corresponds with the displacement of surrounding markers. Local markers in the gel should have roughly similar displacements, and any outliers are likely to be markers which have not converged properly. The displacement of four markers surrounding the current marker (up,down,left,right) are compared in displacement. If there is too much discrepancy between the current marker and an adjacent marker, it is marked as untrustworthy. We determined through testing that marker displacements should not vary more than 15 pixels from each adjacent marker during nominal operation. Finally, an upper limit to the marker displacement is placed to catch any markers that have converged onto different centers. 
After all the markers have been categorized, from the remaining set of trusted markers we interpolate the marker displacements for untrustworthy markers, using linear interpolation. As long the interpolation corrects markers to be near their corresponding markers, the mean-shift algorithm will converge accurately. 

\begin{figure*}[t]
        \centering
        \begin{subfigure}[t]{0.24\textwidth}
            \centering
            \includegraphics[width=\textwidth]{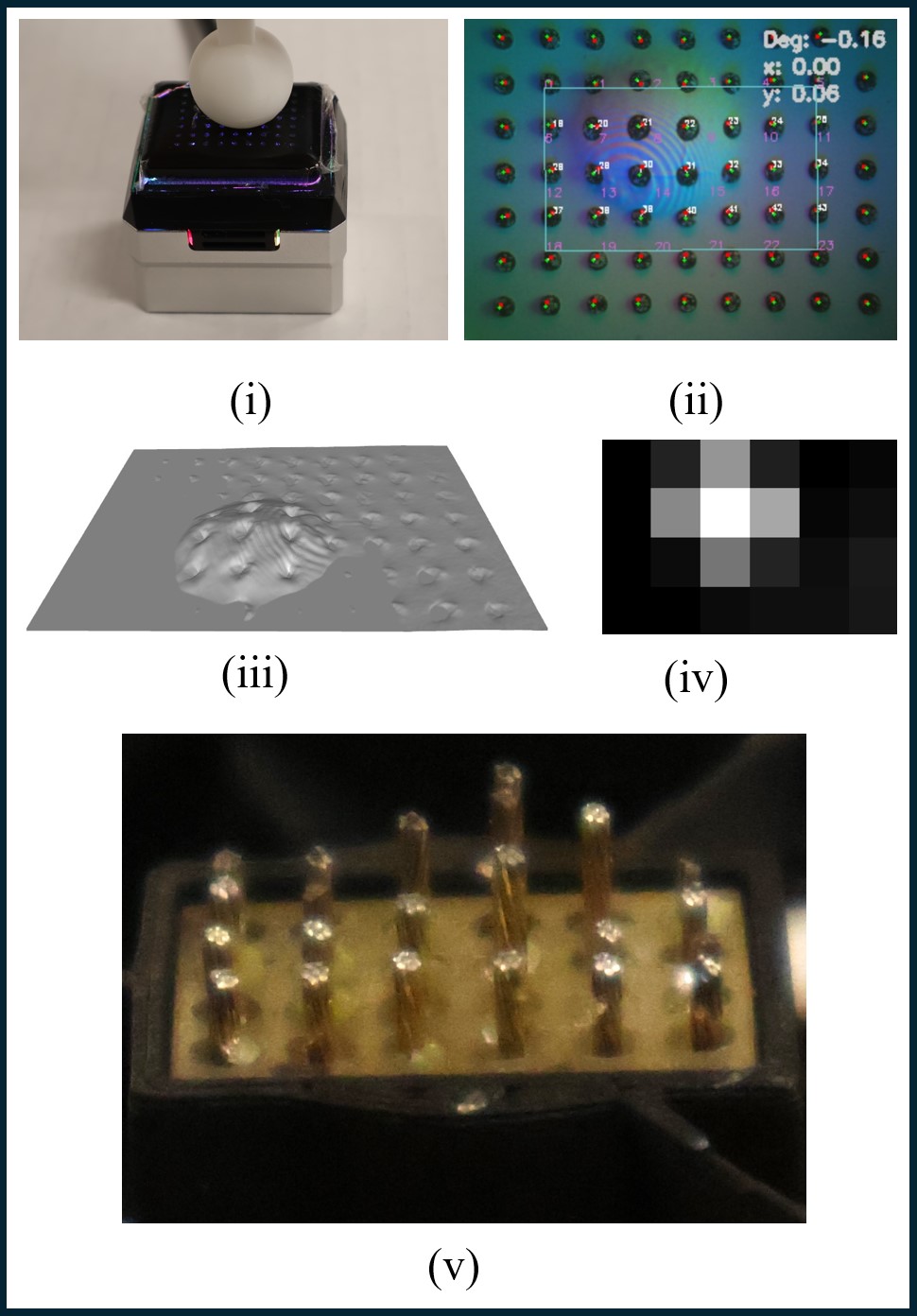}
            \caption{Sphere}
        \end{subfigure}
        \hfill
        \begin{subfigure}[t]{0.24\textwidth}
            \centering
            \includegraphics[width=\textwidth]{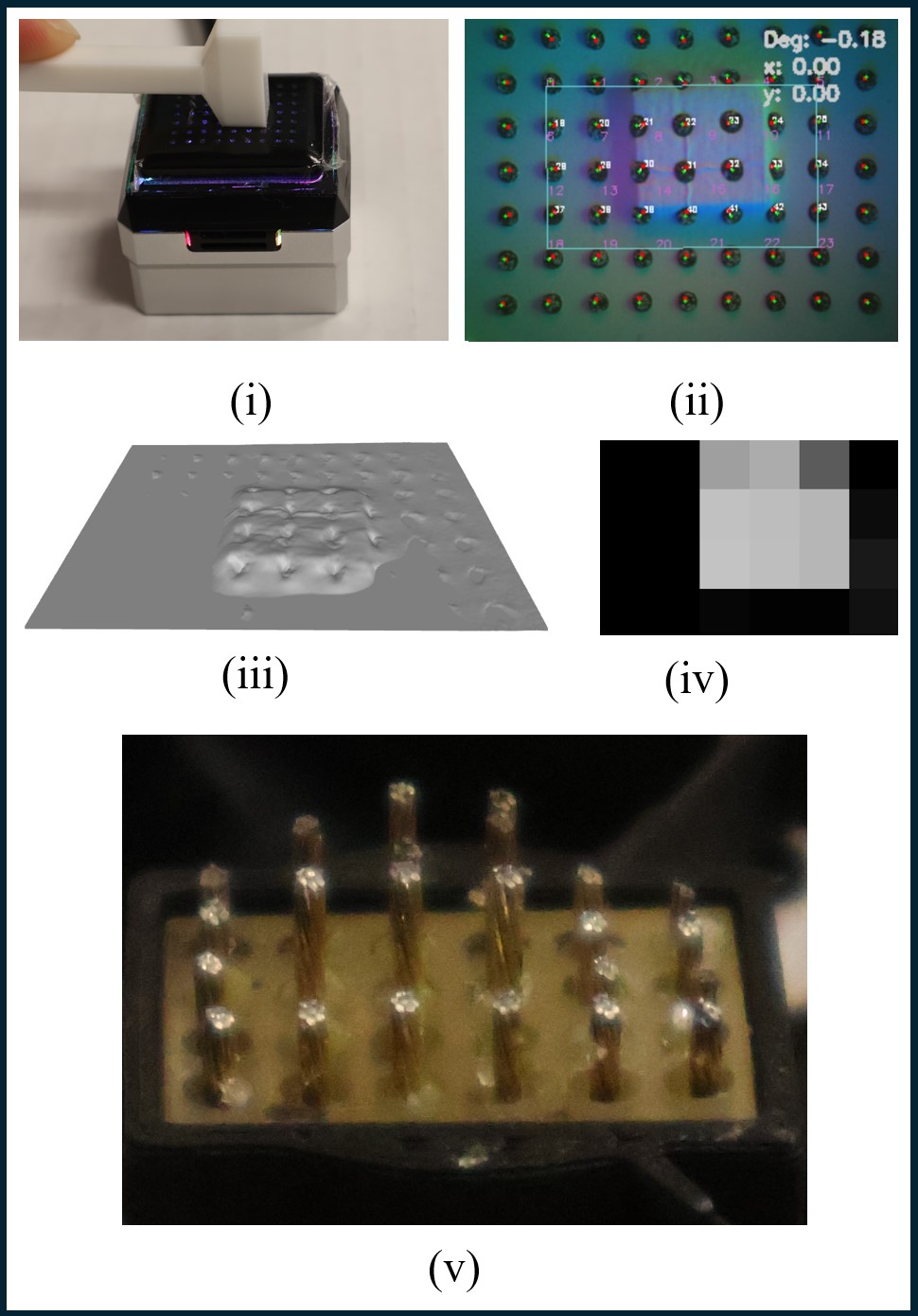}
            \caption{Cube}
        \end{subfigure}
        \hfill
        \begin{subfigure}[t]{0.24\textwidth}
            \centering
            \includegraphics[width=\textwidth]{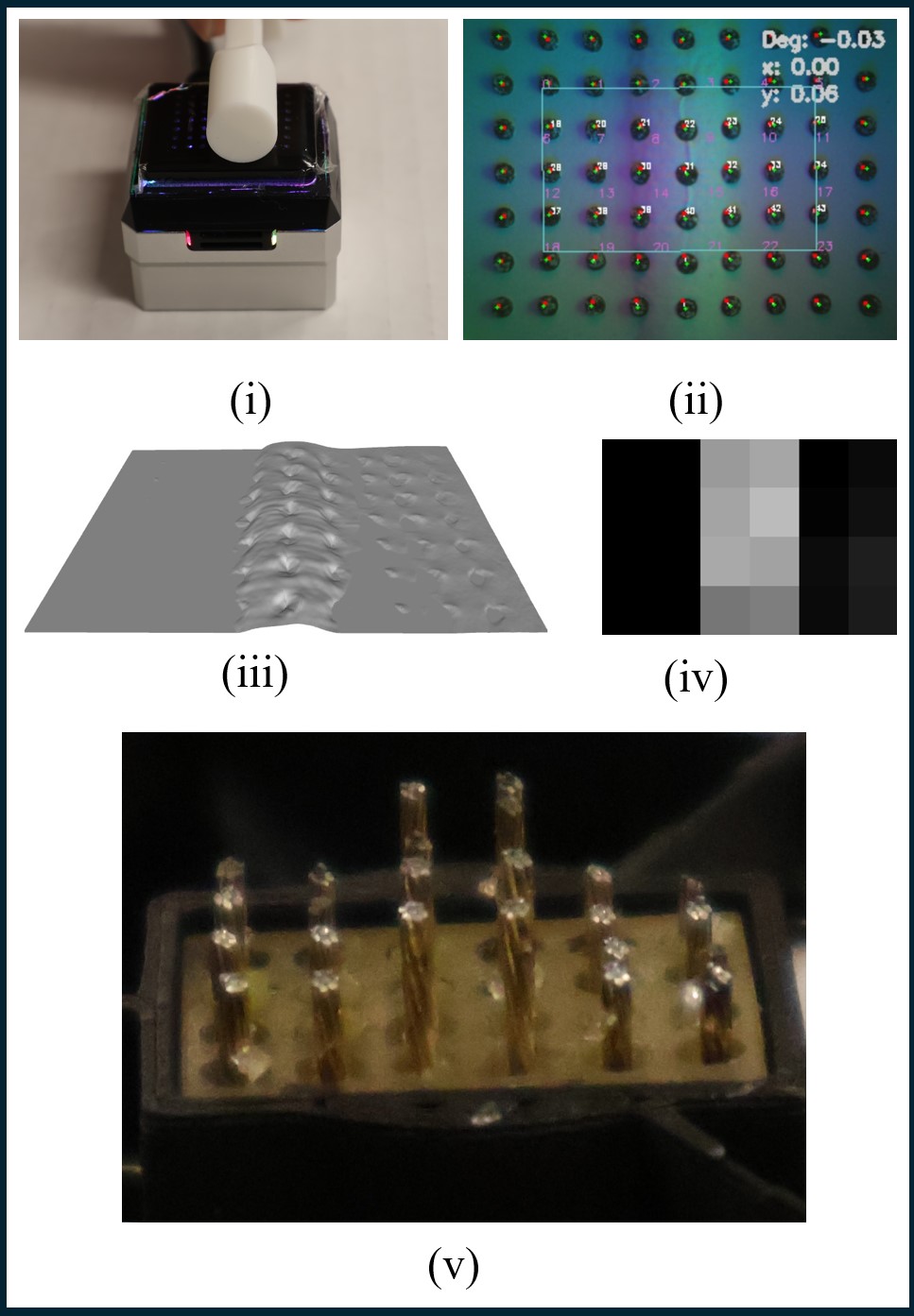}
            \caption{Cylinder}
        \end{subfigure}
        \hfill
        \begin{subfigure}[t]{0.24\textwidth}
            \centering
            \includegraphics[width=\textwidth]{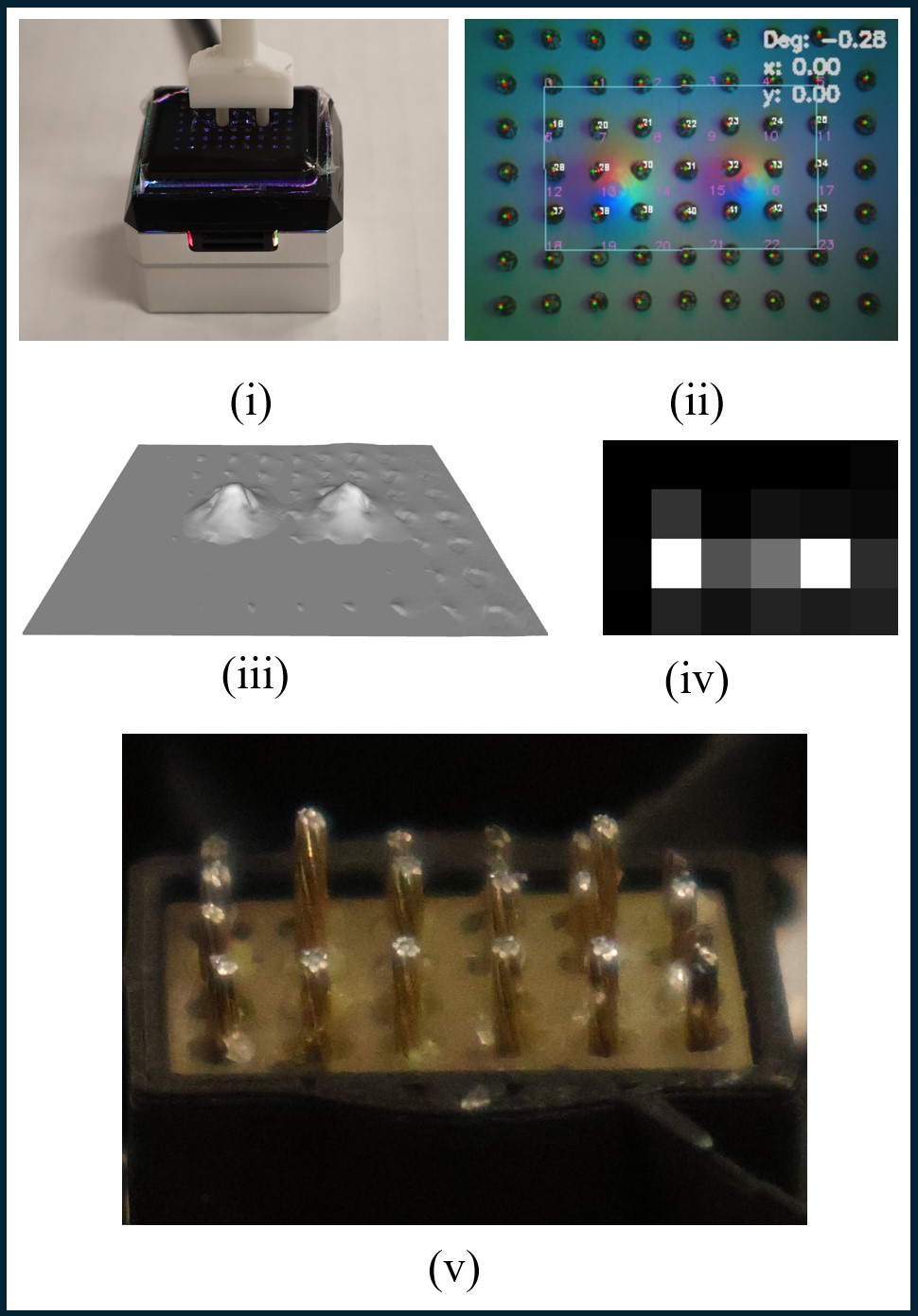}
            \caption{Two Dots}
        \end{subfigure}
          
        \caption{In each subfigure: (i) The object deforming the sensor gel, (ii) raw images of the GelSight sensor, (iii) calculated depth map, (iv) sampled depth map, and (v) physical reconstruction on the pin display face without the gel. The white rectangle in the raw image corresponds to the sampling grid area.}
        \label{fig:depthreconstruction}
\end{figure*}

\begin{figure}[t]
      \centering
      \begin{subfigure}[t]{0.17\textwidth}
            \centering
            \includegraphics[width=\textwidth]{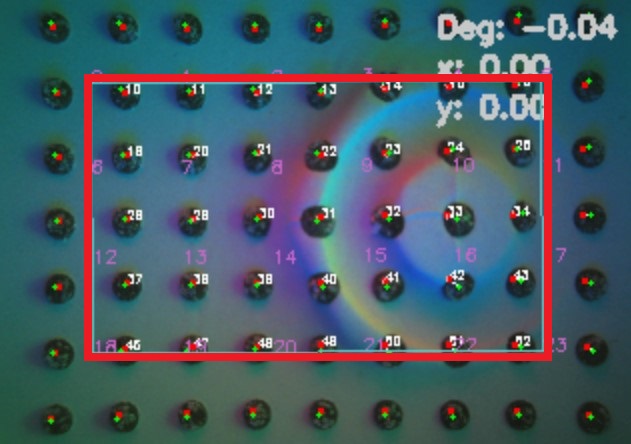}
      \end{subfigure}
      \hfill
      \begin{subfigure}[t]{0.18\textwidth}
            \centering
            \includegraphics[width=\textwidth]{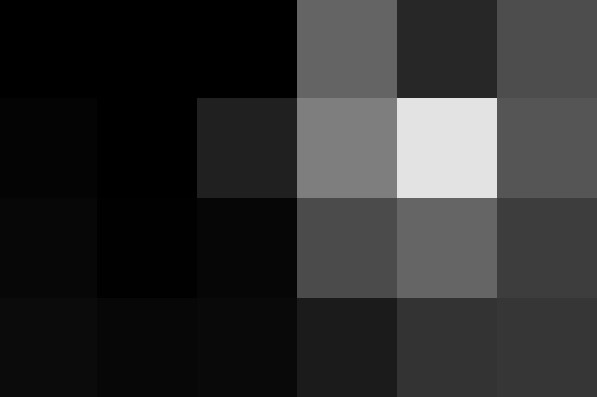}
      \end{subfigure}
      \begin{subfigure}[t]{0.17\textwidth}
            \centering
            \includegraphics[width=\textwidth]{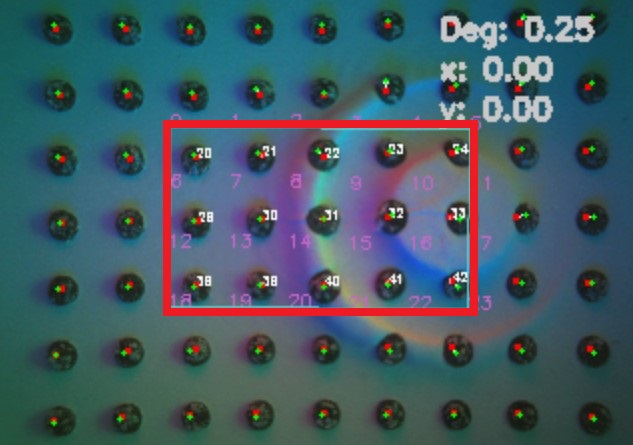}
      \end{subfigure}
      \hfill
      \begin{subfigure}[t]{0.18\textwidth}
            \centering
            \includegraphics[width=\textwidth]{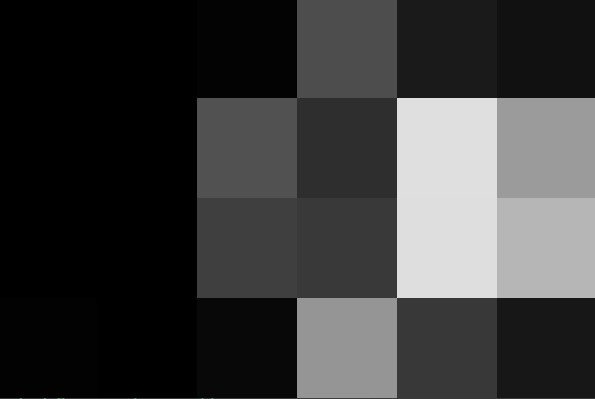}
      \end{subfigure}
      \begin{subfigure}[t]{0.17\textwidth}
            \centering
            \includegraphics[width=\textwidth]{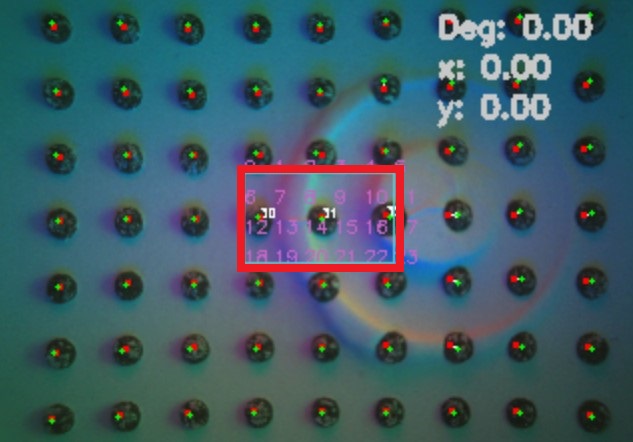}
      \end{subfigure}
      \hfill
      \begin{subfigure}[t]{0.18\textwidth}
            \centering
            \includegraphics[width=\textwidth]{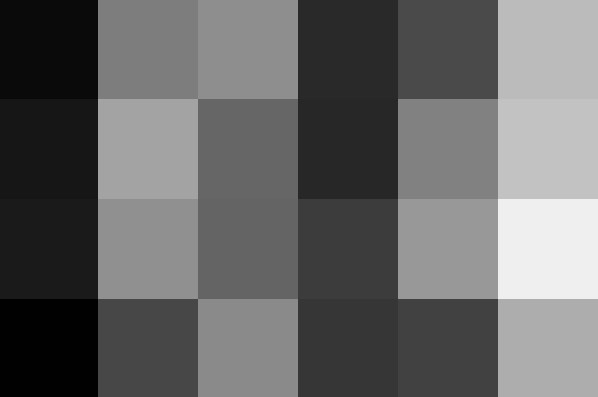}
      \end{subfigure}
      
      \caption{The user can specify the location and spacing of the sampling grid, in order to resolve higher levels of detail. From top to bottom, the sampling spacing is 45, 30, and 15 pixels.}
\end{figure}

\subsection{Sampling \& Reconstruction}

After marker tracking and correction has been applied, a mask created and used to correct the depth map. The tracked markers are then used to calculate the total displacement and rotation of the sampling grid. This sampling grid is used as coordinates to sample the depth map, which correspond to the actuators of each pin on the pin display. The user has the option of specifying the location and size of the grid to be sampled, as well as the gain for sampling. This allows for scaling the sensitivity of the haptic response, which is useful for sensing minute features or grasping small objects. The sampled depths are then translated to serial commands, and send via serial USB to the Pololu Maestro servo controller, as mentioned earlier. The software loop runs at a frequency of ~8 Hz during nominal operation.

\subsection{Device Demonstration}

We demonstrate the operation of our-pin based shape display with the GelSight using several different objects, as seen in Fig. \textcolor{red}{6}. The shapes we tested were a sphere (a), cube (b), cylinder (c), and two dots (d), using 3D printed shapes. The color images are the raw video output from the GelSight sensor, with the black dots being markers for tracking. Overlaid on the image is a white square representing the sampling area, vectors representing the displacement of the markers from the center position, and values for the total X,Y, and rotational displacement calculated. The depth map of the elastomer surface deformed by the object is shown as a 3D point cloud. The 6 by 4 reconstructed depth is also shown as a grayscale image, with longer pin lengths on the pin display represented as whiter colors. Finally, the reconstructed depth and shear is shown on the pin display without the gel face. Due to their relatively large size, marker artifacts are still present in the depth image, even with the mask applied.

We also demonstrate the ability to resolve different levels of detail from the depth image by scaling the sampling area. In Fig. \textcolor{red}{7}, we show 3 different levels of magnification at different locations on the sensing area for an object, by adjusting the spacing in the sampling grid. At high pixel spacing for the sampling grid, general features of the object are apparent, and as the pixel spacing decreases, finer details start apperating, such as the gap between the two concentric circles.

\section{Psychophysics and Teleoperation Experiments}

We now investigate the performance and utility of Feelit through different haptic experiments with human participants \textbf{(IRB-2024-433)}. First, we perform psychophysical experiments without the teleoperation aspect to measure how well humans are able to interpret the tactile information provided by our device. These experiments consists of recognition of simple shapes, shear displacement stimuli, and depth discrimination. Second, we demonstrate the utility of rich tactile feedback with a simple teleoperation task.
For the experiments set, we recruited 7 participants (mean age 24.7, SD=3.4, 2 identifying as female and 5 as male). At the end of each task participants are debriefed on their results and asked about their thoughts on the device. For all experiments, the sampling area and gain were scaled 1-1 with the reconstruction on Feelit.

\subsection{Psychophysics Experiment: Simple Shape Presentation}

The goal of this task is to evaluate how well participants are able to discern simple tactile stimuli that are sensed by the GelSight and reconstructed on our device. Seven simple shapes are impressed on the GelSight by the experimenter and reconstructed on the device as seen in Fig. \textcolor{red}{[8]}. These shapes are a horizontal bar, a vertical bar, two horizontal bars, a diagonal bar, two dots, three dots, and four dots. Shape information is electronically recorded and played back for consistency. In the tutorial phase, participants are presented with each shape at least once, and are allowed to repeat any shape they wish for any duration. Once the tutorial phase ended, participants are blindfolded and are randomly presented with a shape one at a time for 3 seconds. Participants could request to be re-presented with the current shape as many times as they wished. Once participants gave a prediction, their responses were recorded and were presented with a new shape. Participants were not told whether they had determined the correct shape or not. This process is repeated for each shape 3 total times, for a total of 21 presentations per participants. 

\begin{figure}[t]
    \centering
    \includegraphics[scale=0.5]{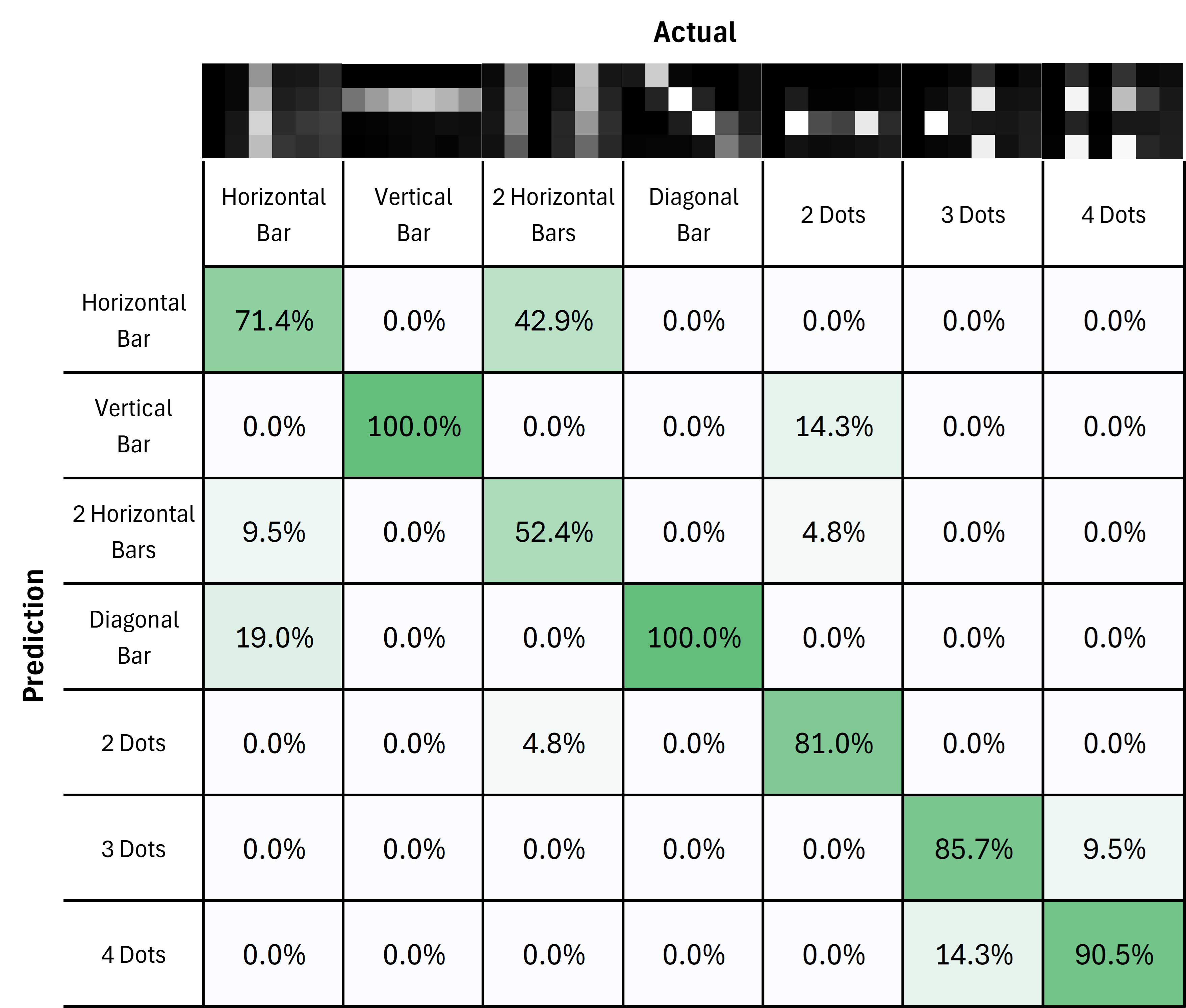}
    \caption{Experiment A: shape presentation results.}
\end{figure}

\subsection{Psychophysics Experiment: Simple Displacement Recognition}

The goal of this task is to evaluate how well participants are able to discern displacement information that are sensed by the GelSight and reconstructed on our compliant mechanism. 6 distinct displacement stimuli are defined for this experiment, and can be seen in Fig. \textcolor{red}{[9]}, which are simulated directly from the software. These stimuli are the display face moving up (distal) from the participant, down (proximal), left, right, and rotating clockwise and counterclockwise. The experiment is conducted in the same manner as in the first experiment. Again, the stimuli are presented 3 times each randomly for a total of 18 presentations per participant. 

\subsection{Psychophysics Experiment: Object Depth Discrimination.}

The goal of this task is to evaluate how well participants are able to differentiate between different depth displacements of an object reconstructed on the display. A sphere is impressed onto the GelSight at a depth of 0.5, 1.0, 1.5, and 2.0 mm, and can be seen in Fig. \textcolor{red}{[10]}, with the response recorded in software for repeatability. The experiment is conducted in the same manner as in the first experiment, with participants receiving a tutorial phase to familiarize themselves. For the experiment phase, the participants will be shown two of the stimuli randomly, with the display being reset for 1 second in between, and have to determine which has a larger depth or if they are the same. Each combination of stimuli are shown twice randomly throughout the experiment, for a total of 20 presentations.

\begin{figure}[t]
    \centering
    \includegraphics[scale=0.6]{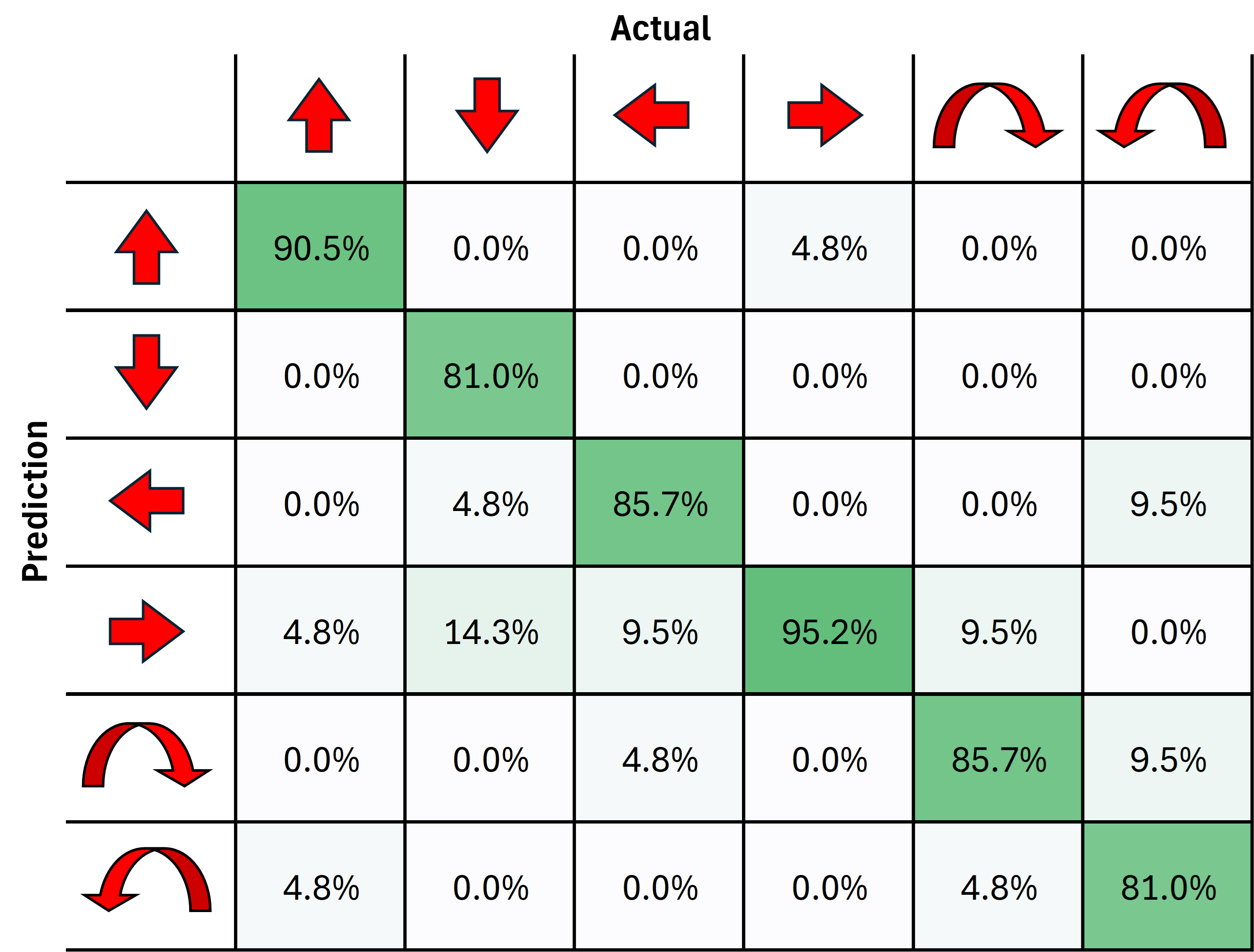}
    \caption{Experiment B: displacement recognition results.}
\end{figure}

\subsection{Teleoperation Demonstration: Relative Weight Identification Task}

The goal of this task is to demonstrate the value of the tactile feedback provided by Feelit in determining object properties. Participants will be operating the Aloha Teleoperation system with the Feelit device attached. Participants will be handed three cylinders identical in appearance, of which two have weights of 250g and 500g embedded inside, and are asked to grasp each with the Aloha robot for a short period of time. We implemented this procedure to eliminate any bias from the participant's individual skill when operating the Aloha device, and any visual or auditory cues that may tip off participants. An initial trial period will be given to familiarize participants with the teleoperation system and task. Once the experiment begins, participants are randomly handed each object one at a time to grasp with the teleoperation system. Participants are instructed to verbally indicate which objects they believe to be the lightest and heaviest, and their level of confidence, with a 0\% confidence indicating a complete guess. The task will be repeated for 5 trials, with the order of the objects randomized between them.

\subsection{Results}

For the shape recognition experiment, participants were able to discriminate the simple shapes with an average accuracy of 83\%, as seen in Fig. \textcolor{red}{[8]}. Although all participants were able to sense the display face, we observed that those with smaller fingers were more confident about their answers as they could move their fingers around the sensing area. This was especially true for the double horizontal bar, which requires the widest sensing area out of all the shapes. Participants commented although they could feel the forward most bar in the double horizontal bar shape, the rear bar was more subtle, which possibly contributed to the confusion between the two. This is supported by the fact that the 2 horizontal bars were misidentified as the single horizontal bar more often than the reverse. Participants were quick to identify distinct shapes such as the diagonal bar or the three dots, but more often than not required a re-presentation for similar shapes such as the 2 dots and vertical bar, or the single and double horizontal bars. Since the gel surface of the pin display provides some form of `smoothing' or `interpolation' between adjacent pins, this may have confused participants in identifying the 2 dots for the vertical bar, but not the other way around.

Participants also performed well on the shear displacement recognition, with an average accuracy of 86.5\% as seen in Fig. \textcolor{red}{[9]}. Notably, since rotation of the display required more servo actuation, some participants were able to discern lateral movement from rotation based on the magnitude of vibration from the servos, even with hearing protection. They were still able to discern clockwise and counterclockwise rotation using tactile sensing only, which is independent of vibrational cues. Participants were overall more confident in predicting shear displacement than the shapes. Participants seemed to be slightly biased towards the right displacement stimulus, which can be seen in the relatively higher amount of misidentifications. A possible explanation is that all participants were right-handed and used the device as such, which may somehow affect decision making.

\begin{figure}[t]
    \centering
    \includegraphics[scale=0.5]{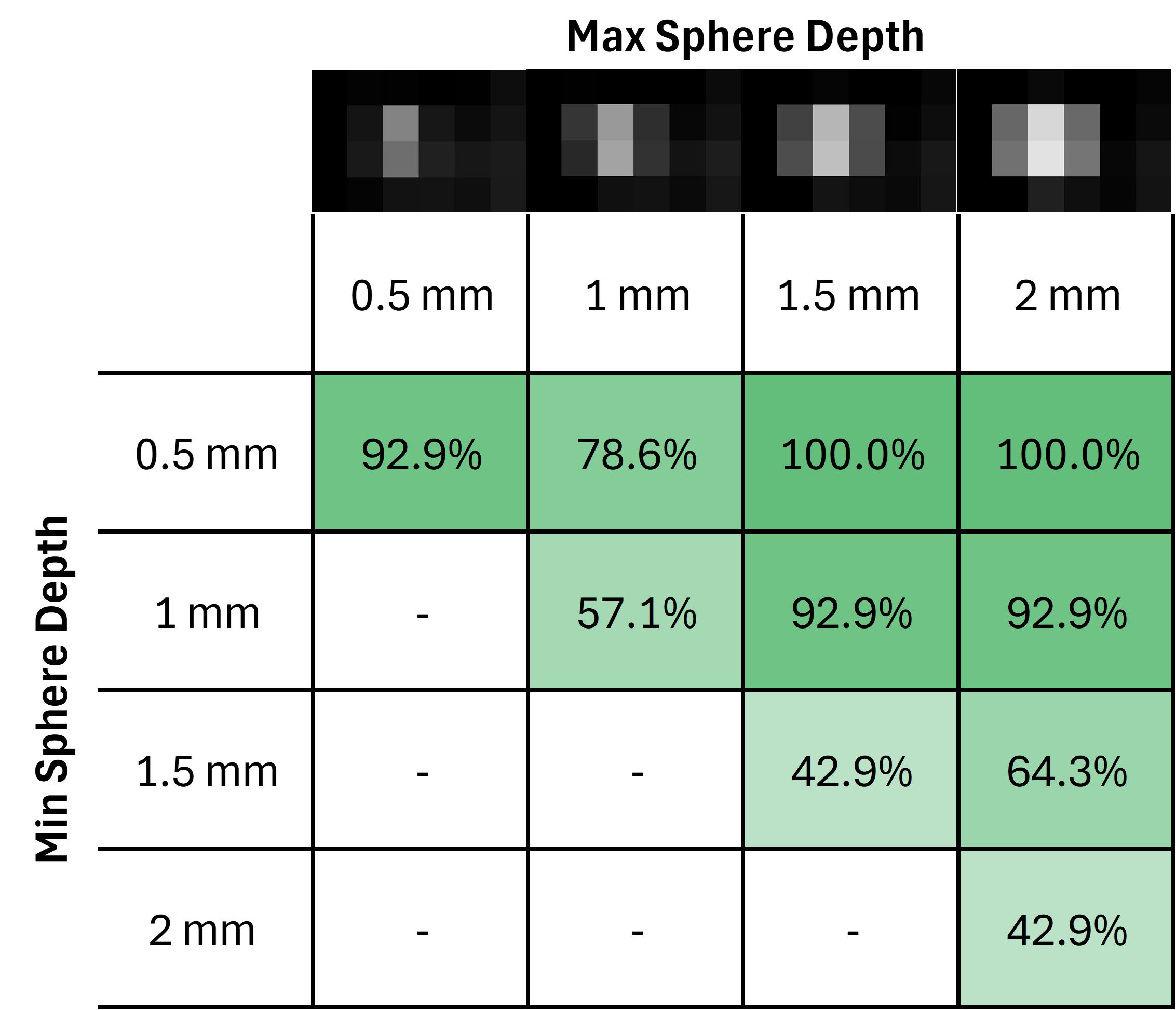}
    \caption{Experiment C: object depth discrimination results. Each cell represents answer accuracy in discriminating between the maximum (column) and minimum (row) depth displacement stimuli pairs}
\end{figure}

\begin{figure}[t]
    \centering
    \includegraphics[scale=0.6]{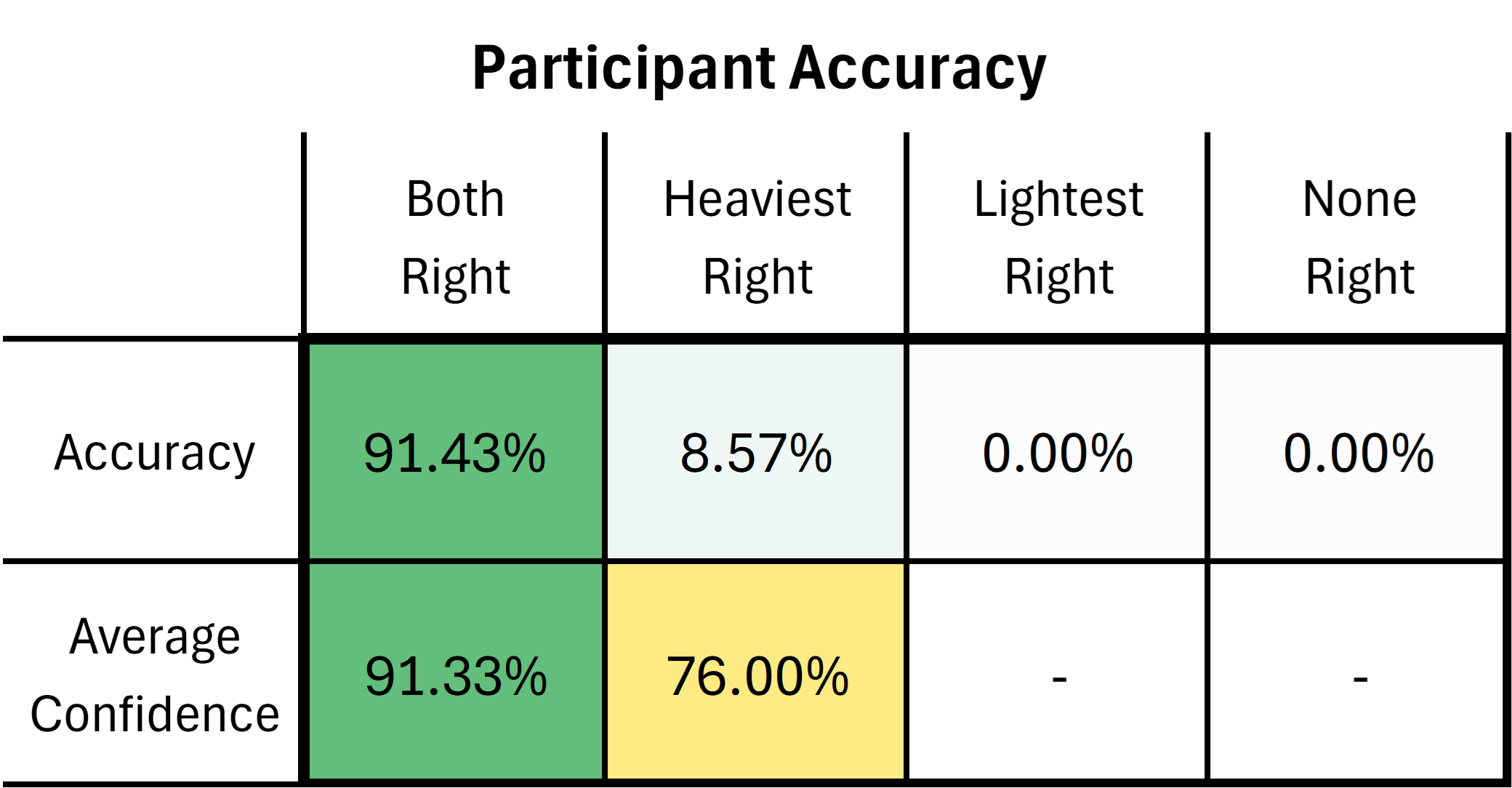}
    \caption{Experiment D: relative weight identification task results. No confidence scores were given for the last two answers since no data was observed}
\end{figure}

For the object depth discrimination (Fig. \textcolor{red}{[10]}), participants were able to discern the difference between spheres with a 1.5 mm difference 100\% of the time. With a 1 mm difference, discrimination accuracy drops to 96.4\%, and with only a 0.5 mm difference, 78.6\%. For sphere presentations with identical heights, participants were not able tell if the heights were the same or different, except in the 0.5 mm high sphere which showed 92.9\% accuracy in identification.

For the relative weight identification task (Fig. \textcolor{red}{[11]}), participants were 91.43\% accurate in identifying the correct relative weights. In trials with both answers right, participants were 91.33\% confident in their answer. Participants only confused the medium and lightest objects 8.57\% of the time while still correctly identifying the heaviest object for all trials. For trials where the lightest object was misidentified, participants had a lower confidence in their answer at 76\%. This suggests logically that there is a correlation between the confidence of a participant's answer and how accurate they were, although more data is needed to evaluate this hypothesis, as participants were overwhelmingly able to identify the correct weights. The test was also conducted for all participants with Feelit disabled. In those tests participants were not at all able or confident in determining the correct relative weights. This, and the high confidence of participant's answers during correct identification, shows that participants were able to utilize the tactile feedback of Feelit in their decision making to determine the object relative weights.

\section{Limitations \& Future Work}

The most apparent limitation of our design is the update frequency of our software loop, due to the heavy computational cost of the depth estimation algorithm, and in part the marker tracking \& error correction. Possible solutions to optimize this are to train a smaller neural network for depth estimation, or estimating tactile response based on the difference image only. Improving the marker tracking \& error correction algorithm or using less markers on the GelSight pad can also speed up computation.

For the physical hardware, although Feelit was small enough to maneuver the leader arm without much difficulty, the servo mechanism housing on the bottom may impede movements when positioning the end effector near the base or low to the ground. Of course, making the device smaller and lighter is another avenue of improvement. Designs to mount the servo mechanisms on the robot links themselves or on opposing sides to balance the weight were considered, but the current design was chosen to balance portability and ease of maintenance, as it is a working prototype. Potential solutions include utilizing smaller actuators or mechanisms, like piezoelectrics. The limited capabilities of the Aloha Teleoperation system also constrained the type of teleoperation experiments conducted. For example, the relatively low precision of Aloha made it difficult for participants to grasp objects in a repeatable manner, and the low actuation force of the gripper made it hard to grasp heavy objects. These limitations resulted in us changing our experimental procedures to maintain consistency. The device, although tailored for use with the Aloha system, can easily be redesigned to fit other similar teleoperation setups that may offer better capabilities. We plan to continue performing more teleoperation tasks to further evaluate the value of the tactile feedback given by our device.

Increasing the resolution of the pin display is also another avenue of advancement. We are reaching the physical limit for pin density with Bowden Cables, but still have yet to utilize the precision offered by the GelSight sensor. Increasing pin density without sacrificing other capabilities like depth resolution and actuation force remains an open research problem.  

\section{Conclusion}

In this paper we present Feelit, our teletaction device that can provide haptic depth reconstruction and shear displacement information to the user. We use this device in conjunction with the GelSight vision-based tactile sensor to relay high-quality tactile feedback to the user in real-time. We accomplish this through miniaturizing a pin-based shape display with Bowden cables and low-cost actuators, and by employing a compliant mechanism to move the display face. Through psychophysics experiments and a relative weight identification task for teleoperation, we show that Feelit is able to effectively convey high quality tactile information to the user in real-time, and has the potential to expand the teleoperation space.

\balance










\bibliographystyle{IEEEtran}
\bibliography{root.bib}

\end{document}